\documentclass[conference,letter]{IEEEtran}

\usepackage{pifont}
\usepackage{pgfplots}
\usepackage{amsthm}
\usepackage{formatting/shortcuts}
\usepackage{xspace}
\usepackage{mathrsfs}
\usepackage{cite}
\usepackage{url}
\usepackage{listings}
\usepackage{boxedminipage}
\usepackage{times}
\usepackage{epstopdf}
\usepackage{array}
\usepackage{enumerate}
 \usepackage{graphicx}
 \usepackage{multirow}
\usepackage{graphicx}
 \usepackage{epstopdf}
 \usepackage{subcaption}
\usepackage{mdwlist}
\usepackage{algorithm}
\usepackage{algpseudocode}
\usepackage{amsmath}
\usepackage{tikzscale}
\usepackage{adjustbox}
\usepackage{booktabs}
\usepackage{hyperref}
\usepackage{enumitem}

\author{Bijeeta Pal  \\ Cornell University \and  Shruti Tople \\Microsoft Research}

\newcommand{\glove}{{GloVe}}
\newcommand{\bert}{{BERT}}

\begin{document}
\title{\huge{To Transfer or Not to Transfer: Misclassification Attacks Against Transfer Learned Text Classifiers} }
\maketitle

\begin{abstract}
Transfer learning --- transferring learned knowledge --- has brought a paradigm shift in the
way models are trained. The lucrative benefits of improved accuracy and reduced training time
have shown promise in training models with constrained computational resources and fewer training samples.
Specifically, publicly available text-based models such
 as \glove{} and \bert{} that are trained on large corpus of datasets have seen ubiquitous adoption in practice. However, the risks involved in
using these public models for various downstream tasks are yet unknown. In this paper, we
ask, {\em ``can transfer learning in text prediction models be exploited to perform misclassification attacks?'' }

As our main contribution, we present novel attack techniques that utilize  
 {\em unintended} features learnt in the teacher (public) model to generate adversarial examples for student (downstream) models. 
 To the best of our knowledge, ours is the first work to show that  transfer learning
from state-of-the-art word-based (e.g., \glove) and sentence-based (e.g., \bert) 
teacher models increase the susceptibility of student models to misclassification attacks.
First, we propose a novel word-score based attack algorithm for generating adversarial
examples against student models trained using context-free word-level embedding model.
On binary classification tasks trained using the  \glove{} teacher model, we achieve an average attack accuracy of $97\%$ for the IMDB Movie Reviews and $80\%$ for the Fake News Detection.
For multi-class tasks, we divide the Newsgroup dataset into 6 and 20 classes and achieve an average
attack accuracy of  $75\%$ and $41\%$ respectively.
Next, we present length-based and sentence-based misclassification
attacks  for the Fake News Detection task trained using a
 context-aware  \bert{} model and achieve  $78\%$ and $39\%$ attack accuracy respectively.
Finally, we observe that existing defenses such as fine-tuning, dropouts and adversarial training fail to mitigate our  attacks. Thus, our results motivate the need for designing
training techniques that are robust to unintended feature learning, specifically for transfer learned models.
 
%
%
%
%
%
%
%

\end{abstract}


\section{Introduction}

Transfer learning techniques use {\em teacher} models pre-trained on large datasets to train {\em student} models for downstream tasks that have smaller amount of data. The idea is to share or transfer the parameters learned in the teacher model and directly use them while training the student model. This avoids training the student model from scratch, i.e., random initialization of parameters, thereby
achieving better model accuracy and reduced training time. This has resulted in transfer learned models being deployed in a variety of different applications~\cite{howard2018universal, huh2016makes,zamir2018taskonomy} and enabled applications that were not possible before~\cite{zhang2016deep, gao2018deep}.
To promote transfer learning for text prediction tasks, state-of-the-art text embedding models like Global Vectors (\glove)~\cite{pennington2014glove} and
Bidirectional Encoder Representations from Transformers (BERT)~\cite{devlin2018bert} are made publicly available. These models are supported  even in popular machine learning frameworks like PyTorch and Tensorflow~\cite{paszke2017automatic,abadi2016tensorflow}. Often, a single teacher model, that has learned generic features of text data, is used for training numerous downstream text prediction tasks. However, the risks involved with such transferring of parameters and re-using features across different models are unknown.  We explore this problem with regards to adversarial examples for transfer learned text-based classification models.

Adversarial examples are designed to fool a classifier into misclassifying them to a specific target class. Prior work on adversarial attacks for text-based classifiers 
 consider either a completely white-box~\cite{lei2019discrete},\cite{li2018textbugger},\cite{liang2017deep} or a black box model~\cite{alzantot2018generating},\cite{gao2018black}.
 However, there has been less or no effort in analyzing 
 the risks of transfer learning  for text data, that provides a part white-box (teacher) and a part black-box (student) setting.
 Recently, researchers have shown misclassification attacks against transfer learned models, but those are limited to only image classification tasks ~\cite{wang2018great,shafahi2018poison,yao2019latent}. 
 Existing techniques such as optimizing the input space
 to find adversarial examples, that work well with images, are ineffective in the text domain due its discrete nature~\cite{papernot2016crafting},\cite{samanta2017towards}. In this paper, we ask, {\em ``can transfer learning in text prediction models be exploited to perform misclassification attacks?'' }
 Our key insight is to leverage {\em unintended} features of inputs to craft adversarial examples
 that get misclassified to a specific output class. Unintended features are not necessarily related to the prediction task
 but are unintentionally learned and used by the student model, as an artifact of tranfer learning from the teacher model or a bias in
 the training dataset. For example,  ``length'' of an article becomes an unintended feature when used for the detection of fake news.

\paragraph{Our Approach.}
To perform misclassification attacks, it is crucial to identify the classification boundary of the output classes in the input feature space. Once this boundary is known, an adversary can carefully craft inputs that belong to one class but cross the boundary in the feature space and get misclassified to a target class. We propose efficient ways to accurately identify these classification boundaries with {\em restricted} access to the student model and {\em no access} to its training dataset. The use of transfer learning opens up a unique threat model where  the teacher part of the model is white-box while the rest (student)  is black-box, we call it the grey-box setting.
 With limited queries to the student model accompanied with unlimited access to the public teacher model, we identify nearly accurate classification boundaries in the feature space. 
We use this boundary as a proxy for the student model and craft adversarial examples in an efficient manner.  We present concrete adversarial attacks for transfer learning from two types of text embedding models --- word-level embedding and sentence-level embedding models.



The word-based embedding model gives a vector representation of words which captures their syntactic and semantic structure. 
Our attack on this model leverages the phenomenon that the prediction probability of a particular class depends on the presence or absence of certain (unintended) features (or words) in the input. Although this phenomenon has been implicitly observed in prior work~\cite{lei2019discrete,alzantot2018generating}, we give a novel and precise scoring mechanism that accurately quantifies the importance of each word towards the prediction of a particular class. To restrict the access to the student model, we train a shadow model that predicts the score of each word for the tasks of the student model.  Thus, in our attack, the number of queries to the student model are independent of the number of adversarial examples to be generated. In contrast, prior work on non-transfer learned models require queries proportional to the number of words in each adversarial input~\cite{alzantot2018generating,li2018textbugger}. We combine the word-scores with the knowledge of the teacher embedding model to construct highly effective targeted adversarial examples. Our adversarial examples preserve semantic correctness, require minimal perturbation and hence are indistinguishable to humans.
The drawback of the word-embedding models is that they do not take the context of the sentence into account. Therefore, we consider a context-aware sentence-based embedding model that represents the entire sentence while capturing its context to a vector. With regards to this model, we present two types of attacks 1) length-based and 2) sentence-based misclassification attacks. 
For these attacks, we use the length and the presence of specific sentences as unintended features to identify the classification boundary in the student model.
Our novel and unconventional approach of using unintended features to create adversarial examples make our attacks robust to existing trivial defenses such as spell-check, grammar check and even more advanced training time defenses such as fine-tuning, dropout or adversarial training.

\paragraph{Results.} 
We evaluate our attacks on transfer learned models using state-of-the-art word-based embedding, \glove{} model~\cite{pennington2014glove}, and 
sentence-based embedding, \bert{} model~\cite{devlin2018bert}. For the \glove{} based transfer  learning, 
we perform concrete attacks on two binary classification tasks — Movie Reviews~\cite{maas-EtAl:2011:ACL-HLT2011} and Fake
News Detection~\cite{fakenewsdataset}, and a multi-class model for the 
Newsgroup dataset~\cite{20newsdataset} divided into $6$ and $20$ classes. On the binary classification tasks, we achieve an average attack accuracy
of  $97\%$  and $80\%$ respectively.  Our attacks outperform prior work on adversarial examples for non-transfer learned models on these datasets~\cite{lei2019discrete},\cite{li2018textbugger},\cite{liang2017deep},\cite{alzantot2018generating},\cite{gao2018black}.
This shows that  {\bf transfer learning a student model from a public teacher model  increases its susceptibility to misclassification attacks.}
On multi-class models, we observe an average misclassification accuracy of $75\%$ for $6$ classes and $41\%$ for $20$ classes. Thus, 
{\bf performing targeted misclassification attacks become difficult with increase in the number of output classes.} 
 For  transfer learning with \bert{} model, our length-based attack on the Fake News Detection model misclassifies $78\%$ of the real news to fake just by reducing the length of the text. Our sentence-based adversarial examples reduce the detection of fake news by $39\%$. This indicates that even {\bf high-capacity context-aware
teacher models like \bert{} are susceptible to misclassification attacks using unintended features.} Our attacks are equally effective on models trained with known defenses such as fine-tuning the teacher model, adding dropout and adversarial training. The robustness to existing defenses is expected as our attacks rely on the unintended features that the model learns to predict output classes. Hence, {\bf mitigating our attacks require techniques that prevent the model from using unintended features for prediction}. However, we speculate that enforcing such constraints might degrade the accuracy of the model which is not acceptable in practice. 

\paragraph{Contributions} Our contribution are as follows:
\begin{itemize}[leftmargin=*]
\squish
	\item {\em New Attack Techniques.} We propose novel attack techniques to generate adversarial examples based on unintended features for the transfer learning setting. We present a word-score based attack for the word-level embedding  model while a length-based, and a sentence-based attack for the sentence-level embedding model. Our attacks require only a few queries to the victim model to identify the unintended features but require no access during the actual online attack. 
	
	\item {\em Efficacy of Attacks.} Our adversarial examples are effective with respect to the misclassification accuracy and efficient with regards to the time required for generating them. Our experiments on state-of-the-art \glove{} and \bert{} models used for training $3$ different downstream tasks exhibit attack accuracy ranging from $97\%$ to $39\%$ depending on the number of output classes and the type of the teacher model.

	\item {\em Robustness to Defenses.} Our  attacks are robust to trivial defenses such as spell-check, grammar check as well as advanced  defenses such as dropout, fine-tuning and adversarial training. Defending our attacks require the transfer learned models to be robust to unintended features in the input data.
	 
\end{itemize}

\section{Problem \& Background}

Our main goal is to understand the susceptibility of transfer learned models towards different misclassification attacks. We first discuss the motivation for transfer learning and the types of teacher and student models for training text-based classifiers. Next, we describe the threat model that arises due to transferring of parameters from teacher to student models.

\subsection{Text-based Prediction with Transfer Learning}
We consider websites or plug-ins that host text classification models such as a neural-network based fake news detector or an ad-blocker~\cite{fakenewsai,adblocker}. We assume that these models are transfer learned from a publicly available teacher model. Teacher models are usually made public by researchers or industries to benefit the community ~\cite{pennington2014glove, devlin2018bert}. Often, there are fewer teacher models on which several applications build their student models. The main idea is that the teacher model learns some common features essential for predicting various other tasks as well. Depending on the similarity of the target task to the original task, the teacher model is used as follows:

\begin{itemize}[leftmargin=*]
\squish
\item {\em Feature Extractor} - The features extracted or the parameters learnt in the teacher model are directly used or copied
as the initial $K$ out of $N$ layers in the student model. The weights of these $K$ layers are fixed (frozen) and are not updated while training the student model. The ratio of $K$ to $N$ determines a deep-layer feature extractor if the ratio is high or a mid-layer feature extractor in case of a lower ratio. A deep-layered feature extractor is more effective if the student task is similar to the teacher task.

\item {\em Fine Tuned} -  In this case, the parameters copied from the teacher model are retrained or fine-tuned with a small learning rate
while training the student model.  Fine-tuning the full model requires more training time, however, it is helpful if the student model's objective is different than the original task of the teacher model. 
\end{itemize}

\begin{figure}
\centering
\includegraphics[width=0.4\textwidth]{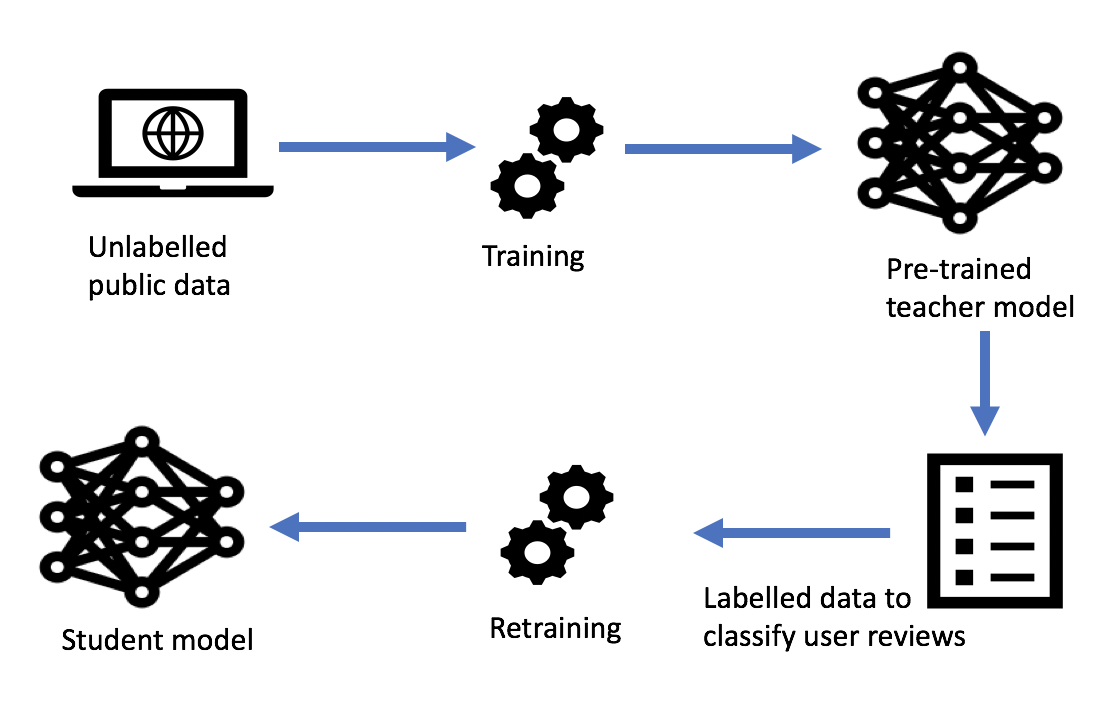}
\caption{Transfer Learning process for text-based classification. }
\label{fig:transfer}
\vspace{-0.5cm}
\end{figure}

Figure~\ref{fig:transfer} shows the process of transfer learning in text prediction models. For text-based models, the teacher models are often called language embedding models as they capture the features in the text data and embed them as vector representations. The initial layer in the student model is initialized using these embedding models and used as a feature extractor or fine-tuned during the training process. There are several teacher models available for text data that provide embedding vectors at the level of characters, words or sentences.

\paragraph{Types of Teacher Models.}
We consider the commonly used word-level embedding and sentence-level embedding which have shown to improve the accuracy of the student models.  We describe both these types below: 

\begin{itemize} [leftmargin=*]
\squish
\item {\em Word-level Embedding: }  Word-level embedding  uses unsupervised learning  to train the teacher model.   It maps  words to vectors such that these vectors preserve the semantic  similarity of the words. Thus, the embedding of similar meaning words have low distance between them.  However, these are context-free models, i.e., each word has a single vector mapped to it irrespective of the context in which it appears for e.g., the word ``orange'' will have the same vector embedding whether it is used as a fruit or the color. \glove{} is a popular word-based embedding model\cite{pennington2014glove}.
\item {\em Sentence-level Embedding:} For these models, the context of the input sentence is considered while training the teacher model. For example, the sentence ``I like to eat orange" specifies the context of orange to make it clear that we are referring to a fruit.
Hence, sentence-level representation performs better than word-level representation. However, they are bulkier, containing millions of parameters to train, making the student model also heavy-weighted. BERT is a state-of-the-art sentence embedding that uses a deep bidirectional transformer encoder to represent the sentence~\cite{devlin2018bert}. 
\end{itemize}

\paragraph{Layers in the Student Model.}
We discuss the layers in the student model that follow the two types of embedding models.

\begin{itemize} [leftmargin=*]
\squish
\item {\em LSTMs:} Long short-term memory (LSTM) model are popular for learning sequence of inputs that have temporal relations between them~\cite{hochreiter1997long}. LSTMs have been widely used in text, speech, audio, and video processing tasks because of its property of remembering patterns for long durations of time. Since word-level embedding model only represents the words, the LSTM layer captures the relationship between the words to perform the text classification task.

\item {\em Linear:} Linear layers are the most simple neural networks with few parameters. They take only fixed-length inputs, making them unsuitable for texts. However, since sentence-level embedding already captures all the contextual information of sentences to a fixed-length vector, the linear layer is appropriate to use after the sentence-level embedding layer.
\end{itemize}

\subsection{Threat Model}

 We consider a passive attacker that has white-box access to the public teacher model but black-box access to the student model. We call this the grey-box setting. 
 The attacker has no access to the training dataset or the student model parameters during or after the training, and hence cannot modify the victim student model. We consider the attacker knows the type of teacher model used i.e., word-level  or sentence-level embedding but has no knowledge about the type of layers (LSTMs, Linear) or parameters in the student model. 
 
The adversary can pose as a genuine user of the application and query the student model to observe the probability distribution of the output classes for any input query.  However, we assume the application has some limit on the number of queries to the model or charges the user per query. 
The attacker can create their own text inputs and submit them to the application. The input text can use any valid English word and printable characters. There is no restriction on the size of the input text or the sentence structure. We consider this attack setting to be realistic for any messaging system or social platforms. The goal of the attacker is to misclassify certain inputs to a desired target class.
For example, the adversary might want all the fake news reports to get predicted as real news. We consider targeted attacks in this paper i.e., misclassify an input to a specific target class as those attacks are highly incentivized as compared to randomized attacks where inputs are misclassified to any of the output classes.

\subsection{Problem Statement: Misclassification Attacks}

Our goal is to design attack algorithms to generate inputs such that they are misclassified to a desired target class. Before we state our problem of misclassification attack, we  describe the classification procedure in transfer learned models.

\paragraph{Classification with transfer learning.}
Let us consider a text binary classification task, without loss of generality, consisting of input-label pairs $(x,y) \in A \times \{+1,-1\}$ from a distribution $D$. The objective is to learn a classifier $C:  \rightarrow \{+1,-1\}$, that takes an input text $x$ drawn from a distribution representing all valid string inputs $A$ ($x \sim A$) and maps it to a label $y \in \{+1,-1\}$. A feature extractor is defined as a function $f$ that takes a input $x$ from $A$ and maps it to an $n$-dimensional vector $R^n$, $F:A \rightarrow R^n$. 
Then, $(f_w, f_s) \in F$ where $f_w$ and $f_s$ are word-level and sentence-level feature extractors respectively.
A student model thus can be defined as $C= (f,w,b)$ where $w$ is a weight vector and $b$ is a bias value.  Thus for an input $x$, the classifier will then make a prediction based on the dot product between feature $f(x)$ and model weight $w$. 

\begin{equation}
  C(x)=\left\{
  \begin{array}{@{}ll@{}}
    +1, & \text{if}\ \sum_i w_i \cdot f(x) + b > 0 \\
    -1, & \text{otherwise}
  \end{array}\right.
\end{equation} 

The attacker has access to the features $f(.)$ learned form the teacher model that are transferred and used by the text classifier $C$. The attackers objective is to misclassify inputs to a specific output class. We describe these attacks below:
 
\paragraph{Misclassification or Adversarial Attacks.}
Misclassification or adversarial attacks involve performing small perturbations to any input such that the changes are unrecognizable by a human but fools the classifier into predicting the desired output class. Let $\delta$ be any perturbation from a set of valid perturbations $\Delta$. Valid perturbations are the feasible modifications an attacker can make to the input under the specified threat model. For example, adding a sentence conveying the same message in different words is a valid perturbation as an adversary can write any number of sentences. The goal of the adversary is to generate an adversarial input $x'$ for an input $x$ such that

\begin{equation}
\min_\delta C(x') \neq C(x)~where~  x+\delta = x'
\end{equation}

Similar attacks are shown on image classification models~\cite{shafahi2018poison}. However, their techniques do not apply in text domain because of the discrete nature of inputs. Arbitrary perturbation might lose the semantic correctness of the text input making it completely senseless and easily detectable.   In comparison, it is easier to modify images without being distinguished by a human observer.  Often, the perturbations are conditioned such that the $L_2$ distance of the modified image from the original image is less than a threshold $th$ to keep the noise undetectable. Adding such unnoticeable noise is not possible in texts, as changing a single character is easily detectable by spellcheckers\cite{li2018textbugger}. Also, changing a word in a sentence can modify the whole semantic structure of the input. Therefore, accounting for closeness in the semantic meaning of the input is important than the absolute distance of characters or words. Nevertheless, existing literature for adversarial samples in text compute the  amount of perturbation as the fraction of replaced words and bound it using a threshold $th$~\cite{alzantot2018generating},\cite{lei2019discrete}.  We use the same metric and leave the task of finding a justifiable way of defining adversarial examples for text data to future work.

\subsection{Attack overview}


Our attack strategies differ for transfer learning with word-based and sentence-based embedding models. However, all of them are  based on the common theme of models making predictions based on unintended features. 

\paragraph{Unintended Features.}
Neural networks learn features from the training dataset that lead to high accuracy for the validation data but the prediction might infact use some unintended features. 
Recent papers have shown that deep learning models often make predictions based on undesirable features like race, gender and so on~\cite{bolukbasi2016man}. Ilyas et al. show that image classification models, while optimizing features that increase accuracy, can learn from useful but non-robust features\cite{ilyas2019adversarial}. Presence of non-robust features result in the models being susceptible to adversarial perturbations. In this work, we empirically show that  text classification models make predictions based on features that do {\em not} capture the task intended to perform by the model. For example, while certain words like `Wikileaks' and `Sources' are more common in fake news than in real news, classifying a news as fake purely based on the presence of these words is not expected from the model.
It can be argued that if the training dataset contains this bias then maybe it is indeed an important feature for classifying news. This argument is true if the test dataset also follows the same distribution of words. However, the distribution of words is often user-dependent and therefore can be maliciously changed under our threat model. In this paper, we present attacks that exploit these unintended features, either transferred from the teacher model or learned by the student model. 

\paragraph{Attack Algorithms.} Due to the difference in the training process of teacher models, our attack strategies are based on different features. For word-level embedding, the words present in the input and their representation $f_w(w_i)$ plays an important role in the prediction.
The attacks on sentence-level embedding model are dependent on the output value of the embedding representation of input $x$, $f_s(x)$.  We exploit this knowledge in our attack and propose novel strategies to identify the 
classification boundaries for binary and multiple tasks using the observable features from the public teacher model. Essentially, we query the teacher model with a input $x$ to collect the features and query the black-box student model with the same input to observe the output class. We demonstrate that with limited queries to the student model accompanied with those to the teacher model, we identify a classification boundary in the feature space. We use this boundary in the feature space as a proxy for the student model and craft adversarial samples in an efficient manner. Unlike prior work, we do not require to query the student model during the online attack~\cite{alzantot2018generating}. We present a concrete word-score based attack alogrithm in Section~\ref{sec:wembedding_res}, and length and sentence-based attack in Section~\ref{sec:sembedding}.

\section{Word-level Embedding  Attack Details}
 \label{sec:wembedding}
 
 Word-level embedding models were first introduced by Mikolov et al.\cite{mikolov2013distributed}. They represent an input word in the form of an $n$-dimensional real vector.
  The word embeddings are the standard way of representing inputs for training natural language models. For the explanation of our attacks in this section, we use the  \glove{} embedding model as our teacher model to train different student models.
The student model initializes its weights with the \glove{} vectors and is either fine-tuned called \glove$_{\tt FT}$ or used simply as a feature extractor called \glove$_{\tt FE}$.
We consider student model trained for binary and multiple classification tasks. We use the Fake News Detection and Newsgroup dataset divided into $6$ and $20$ classes as our representatives
for binary and multi-class models (details in Section~\ref{sec:wembedding_res}). However, our attack techniques are generic and are applicable to {\em any} text-based classification task transfer learned using {\em any} word-level embedding model.

\subsection{ Word-score based Prediction}

We hypothesize that the student models that are trained using word-level embedding teacher model
predict the output class purely based on the words present in the input sample. This
hypothesis relies on two key characteristics of  language models. First, models memorize words present in the training dataset and use them as important
features to predict the output. Second,  the importance of a word for prediction is directly related to the frequency of that word in the training dataset 
with respect to the specific output class. For example, a common word such as ``The'' that appears frequently in inputs of all the classes will not be an important feature to predict a particular output class.
These observations are in-line with previous work~\cite{li2018textbugger,alzantot2018generating,lei2019discrete} that show importance of a few words in classification of an input. In this
work, we propose a new technique to calculate the importance or {\em score} of each word towards the output class. Using the word-score metric, we propose a method to identify
the classification boundaries for binary and multiple classification tasks. We validate the accuracy of our word-score based prediction and observe it to be close to the original ground truth values.

%
%
%
%
%
%

\paragraph{Word-score for Binary Tasks.}
Let the student model S have two output classes A and B and the word $w$  has the output probabilities as $p_A$ and $p_B$.  Each word has a score $K$ corresponding to each class. Then, the score for $w$ is:
\begin{gather}
o= \arg\max_{x \in \{A, B\}} p_x \\
K_w[o] = p_{o} - p_{o'} \\
K_w[o'] = 0 \label{eqn:einstein}
\end{gather}
where $o'$ represents the class other than $o$. Using this metric, we calculate the score of an input sample I = $w_1 \ldots w_n$
by adding the score of each word in their corresponding class. Thus, the score $K$ of the input $I$ for each class $j$ is: 
\begin{equation}
K_I[j] = \sum_{i=1}^n K_{w_i}[j] 
\end{equation} 
 
Our word-score based prediction method assigns an input $I$ to a class $j$ with the maximum score as follows.
 \begin{equation}
j =  \arg\max_{x \in \{A, B\}} K_I[x]
\end{equation}

The above method gives accurate prediction when the words in the vocabulary are distributed evenly among the output classes i.e, the number of positive scored words in class A are almost equal to those in class B.
 Often, the distribution of words is skewed towards a particular class. Hence, we propose an advanced word-score prediction method that considers the skewness in the distribution of words.
  Let $t_A$ be the ratio of words with respect to all the words in the vocabulary that belong to class A, then 
\begin{equation}
j =  \arg\max_{x \in \{A, B\}} \frac{K_I[x]}{t_x}
\end{equation}

\paragraph{Word-score for Multiple Tasks.}
In a multiple classification task, the model predicts among $n$ different output classes. The overall method for calculating the word-score remains the same as in binary classification except that we consider $n$ classes for scoring each word. We can generalize the scoring Eq. \ref{eqn:einstein} to $n$ classes with each class having probability $p_1 \ldots p_n$. 
\begin{gather}
o = \arg\max_{x \in \{1,\ldots, n\}} p_x \\
K_w[o] = p_{o} - \frac{1}{n-1}\sum_{i = 1, i \neq o}^n p_i \\
K_w[o'] = 0 
\end{gather}
where $o'$ is for all the classes except the one with the maximum score. The final output class is the one with the maximum sum of word-scores compared to other classes. With the increase in number of classes, each class receives a lower total score and the class having the maximum neutral words or common words becomes dominating. To make sure that the word-score indicates the presence of words particular to a specific class, for multi-class classifiers, we include a word score only if it is greater than a minimum threshold of the scores. This threshold is
identified empirically after observing the word-score  for all the words in the vocabulary across all the classes.
 This modified scoring strategy removes the common words that have low scores but are present in majority, and act as noise towards the prediction. This problem is less relevant in binary tasks as the total sum score for each class is high enough to not let the common words impact the classification.

\subsection{Validating Word-score Heuristic}
We evaluate how good is of our word-score heuristic on binary and multiple classification tasks  for Fake News Detection and Newsgroup dataset respectively.

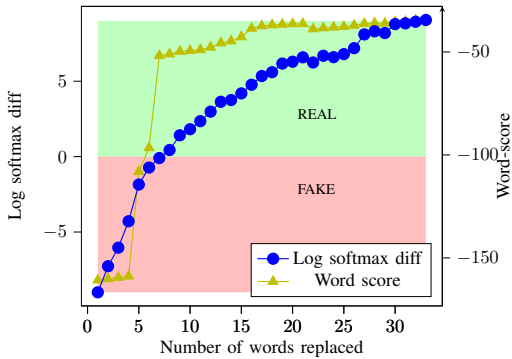
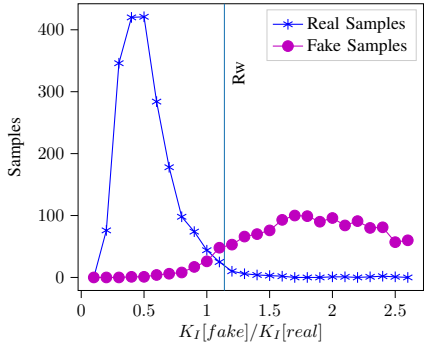
\begin{figure}
	\centering
	\begin{subfigure}[b]{\linewidth}
		\centering
\begin{tikzpicture}[scale=0.7]

\definecolor{color0}{rgb}{0.5,1,0.5}
\definecolor{color1}{rgb}{1,0.5,0.5}
\definecolor{color2}{rgb}{0.75,0.75,0}

\begin{axis}[
tick align=outside,
tick pos=left,
x grid style={white!69.01960784313725!black},
xlabel={Number of words replaced},
xmin=-0.6, xmax=34.6,
xtick style={color=black},
y grid style={white!69.01960784313725!black},
ylabel={Log softmax diff},
ymin=-9.903015, ymax=9.963315,
ytick style={color=black}
]
\path [fill=color0, fill opacity=0.5]
(axis cs:1,9)
--(axis cs:1,0)
--(axis cs:2,0)
--(axis cs:3,0)
--(axis cs:4,0)
--(axis cs:5,0)
--(axis cs:6,0)
--(axis cs:7,0)
--(axis cs:8,0)
--(axis cs:9,0)
--(axis cs:10,0)
--(axis cs:11,0)
--(axis cs:12,0)
--(axis cs:13,0)
--(axis cs:14,0)
--(axis cs:15,0)
--(axis cs:16,0)
--(axis cs:17,0)
--(axis cs:18,0)
--(axis cs:19,0)
--(axis cs:20,0)
--(axis cs:21,0)
--(axis cs:22,0)
--(axis cs:23,0)
--(axis cs:24,0)
--(axis cs:25,0)
--(axis cs:26,0)
--(axis cs:27,0)
--(axis cs:28,0)
--(axis cs:29,0)
--(axis cs:30,0)
--(axis cs:31,0)
--(axis cs:32,0)
--(axis cs:33,0)
--(axis cs:33,9)
--(axis cs:33,9)
--(axis cs:32,9)
--(axis cs:31,9)
--(axis cs:30,9)
--(axis cs:29,9)
--(axis cs:28,9)
--(axis cs:27,9)
--(axis cs:26,9)
--(axis cs:25,9)
--(axis cs:24,9)
--(axis cs:23,9)
--(axis cs:22,9)
--(axis cs:21,9)
--(axis cs:20,9)
--(axis cs:19,9)
--(axis cs:18,9)
--(axis cs:17,9)
--(axis cs:16,9)
--(axis cs:15,9)
--(axis cs:14,9)
--(axis cs:13,9)
--(axis cs:12,9)
--(axis cs:11,9)
--(axis cs:10,9)
--(axis cs:9,9)
--(axis cs:8,9)
--(axis cs:7,9)
--(axis cs:6,9)
--(axis cs:5,9)
--(axis cs:4,9)
--(axis cs:3,9)
--(axis cs:2,9)
--(axis cs:1,9)
--cycle;

\path [fill=color1, fill opacity=0.5]
(axis cs:1,-9)
--(axis cs:1,0)
--(axis cs:2,0)
--(axis cs:3,0)
--(axis cs:4,0)
--(axis cs:5,0)
--(axis cs:6,0)
--(axis cs:7,0)
--(axis cs:8,0)
--(axis cs:9,0)
--(axis cs:10,0)
--(axis cs:11,0)
--(axis cs:12,0)
--(axis cs:13,0)
--(axis cs:14,0)
--(axis cs:15,0)
--(axis cs:16,0)
--(axis cs:17,0)
--(axis cs:18,0)
--(axis cs:19,0)
--(axis cs:20,0)
--(axis cs:21,0)
--(axis cs:22,0)
--(axis cs:23,0)
--(axis cs:24,0)
--(axis cs:25,0)
--(axis cs:26,0)
--(axis cs:27,0)
--(axis cs:28,0)
--(axis cs:29,0)
--(axis cs:30,0)
--(axis cs:31,0)
--(axis cs:32,0)
--(axis cs:33,0)
--(axis cs:33,-9)
--(axis cs:33,-9)
--(axis cs:32,-9)
--(axis cs:31,-9)
--(axis cs:30,-9)
--(axis cs:29,-9)
--(axis cs:28,-9)
--(axis cs:27,-9)
--(axis cs:26,-9)
--(axis cs:25,-9)
--(axis cs:24,-9)
--(axis cs:23,-9)
--(axis cs:22,-9)
--(axis cs:21,-9)
--(axis cs:20,-9)
--(axis cs:19,-9)
--(axis cs:18,-9)
--(axis cs:17,-9)
--(axis cs:16,-9)
--(axis cs:15,-9)
--(axis cs:14,-9)
--(axis cs:13,-9)
--(axis cs:12,-9)
--(axis cs:11,-9)
--(axis cs:10,-9)
--(axis cs:9,-9)
--(axis cs:8,-9)
--(axis cs:7,-9)
--(axis cs:6,-9)
--(axis cs:5,-9)
--(axis cs:4,-9)
--(axis cs:3,-9)
--(axis cs:2,-9)
--(axis cs:1,-9)
--cycle;

\addplot [semithick, color2, mark=triangle*, mark size=3, mark options={solid}]
table {%
1 -8.192
2 -8.1144
3 -8.0264
4 -7.9306
5 -0.9958
6 0.5834
7 6.6981
8 6.7989
9 6.972
10 7.0273
11 7.0838
12 7.249
13 7.5329
14 7.6617
15 7.9318
16 8.5058
17 8.6615
18 8.728
19 8.7695
20 8.7988
21 8.8107
22 8.4625
23 8.5312
24 8.5794
25 8.6192
26 8.7035
27 8.8201
28 8.8506
29 8.8369
30 8.9466
31 8.9582
32 9.055
33 9.0603
};\label{plot_one}

\node at (axis cs:20,-2.5)[
  scale=0.8,
  anchor=base west,
  text=black,
  rotate=0.0
]{FAKE};
\node at (axis cs:20,2.5)[
  scale=0.8,,
  anchor=base west,
  text=black,
  rotate=0.0
]{REAL};
\end{axis}

\begin{axis}[
axis y line=right,
tick align=outside,
x grid style={white!69.01960784313725!black},
xmin=-0.6, xmax=34.6,
xtick pos=left,
xtick style={color=black},
y grid style={white!69.01960784313725!black},
ylabel={Word-score},
ymin=-173.3098415, ymax=-27.8713285,
legend pos=south east,
ytick pos=right,
ytick style={color=black}
]

\addplot [semithick, blue, mark=*, mark size=3, mark options={solid}]
table {%
1 -166.699
2 -154.02333
3 -145.01222
4 -132.20694
5 -114.36348
6 -106.122444
7 -101.51904
8 -97.621445
9 -90.46946
10 -87.5242
11 -83.602844
12 -78.99361
13 -74.19005
14 -73.34587
15 -70.1195
16 -65.922585
17 -61.682693
18 -59.838875
19 -55.64614
20 -54.75248
21 -52.619987
22 -55.11477
23 -51.855522
24 -52.531044
25 -51.05574
26 -48.177208
27 -41.48905
28 -39.987526
29 -40.82827
30 -36.493187
31 -36.121437
32 -35.35442
33 -34.48217
}; \label{plot_two}

\addlegendimage{/pgfplots/refstyle=plot_one}\addlegendentry{Log softmax diff}
\addlegendimage{/pgfplots/refstyle=plot_two}\addlegendentry{Word score}

\end{axis}

\end{tikzpicture}
		\vspace{-0.5cm}
		\caption{Word-score is related to the prediction probability of the classifier}
		\label{fig:hypothesis}
	\end{subfigure}

	\begin{subfigure}[b]{\linewidth}
		\centering
		\vspace{0.2cm}
		\begin{tikzpicture}[scale=0.67]

\definecolor{color0}{rgb}{0.75,0,0.75}
\definecolor{color1}{rgb}{0.12156862745098,0.466666666666667,0.705882352941177}

\begin{axis}[
legend cell align={left},
legend style={draw=white!80.0!black},
tick align=outside,
tick pos=left,
x grid style={white!69.01960784313725!black},
xlabel={$K_I[fake]/K_I[real]$},
xmin=-0.025, xmax=2.725,
xtick style={color=black},
y grid style={white!69.01960784313725!black},
ylabel={Samples},
ymin=-21.05, ymax=442.05,
ytick style={color=black}
]
\addplot [semithick, blue, mark=asterisk, mark size=3, mark options={solid}]
table {%
0.1 0
0.2 76
0.3 346
0.4 420
0.5 421
0.6 284
0.7 178
0.8 98
0.9 74
1 44
1.1 25
1.2 10
1.3 6
1.4 4
1.5 3
1.6 2
1.7 0
1.8 0
1.9 0
2 1
2.1 1
2.2 0
2.3 1
2.4 2
2.5 1
2.6 0
};
\addlegendentry{Real Samples}
\addplot [semithick, color0, mark=*, mark size=3, mark options={solid}]
table {%
0.1 0
0.2 0
0.3 0
0.4 1
0.5 1
0.6 4
0.7 6
0.8 8
0.9 17
1 26
1.1 48
1.2 53
1.3 66
1.4 70
1.5 76
1.6 93
1.7 100
1.8 99
1.9 90
2 96
2.1 84
2.2 91
2.3 80
2.4 81
2.5 57
2.6 60
};
\addlegendentry{Fake Samples}
\addplot [semithick, color1, forget plot]
table {%
1.14 -21.05
1.14 442.05
};
\node at (axis cs:1.3,300)[
  scale=1,
  anchor=base west,
  text=black,
  rotate=90.0
]{Rw};
\end{axis}

\end{tikzpicture}
				\vspace{-0.5cm}
		\caption{ Number of misclassified fake and real samples based on different ratios of $\frac{K_I[fake]} {K_I[real]}$. Let $t_f = $ total fake words and $t_r = $ total fake words, then $Rw = \frac{t_f}{t_r}$   }
		\label{fig:word_ratio}
    \end{subfigure}
    \caption{Word-score based heuristics for identifying the classification boundary for Fake News Detection classifier}
    \label{fig:valid_word_score}
    		\vspace{-0.5cm}
\end{figure}

\paragraph{Binary Tasks.}
We aim to validate how close or good is the classification boundary identified using our word-score mechanism. 
First, we take an input text which is rightly classified as a fake news by our binary classifier trained for fake news detection.
Next, we replace the words with a high fake score to words with real score in this input and query the classifier again.
Figure \ref{fig:hypothesis} shows the change in the difference of log probability and word-score with the number of words replaced in the input.
The classification flips at one point and the model predicts the sentence to be real with high probability. This shows that there is a direct correlation with words-score sum and the input sentence classification. 
The exact correlation is identified by observing the distribution of words between the two classes. Figure~\ref{fig:word_ratio} shows the ratio of $K_I[fake]$ and $K_I[real]$, across various real and fake news inputs from the dataset.
Most of the real news  have a ratio of less than 1, indicating a higher value of $K_I[real]$. Similarly,  majority of fake samples have a ratio greater than 1. The $R_w$ is the ratio of the total number of fake words divided by the total number of real words in the entire vocabulary. Thus, $R_w$ is a good estimate for classifying if a sample is real or fake. 

We compare the output class predicted using our word-score mechanism to the ground truth values of $2000$ Real and $2000$ Fake News samples.
 The word-scores are calculated by querying the student model that is trained using both \glove$_{\tt FT}$ and \glove$_{\tt FE}$.
 Table \ref{tabel:transfer_learning} shows that our word-score based mechanism identifies the prediction boundary with $97.4\%$ and
$85.82\%$ accuracy when the student model is trained with  \glove$_{\tt FT}$ and \glove$_{\tt FE}$ respectively. The student model itself
reaches $95.4$\% and $95.1\%$ accuracy as shown in Section~\ref{sec:wembedding_res}.
This shows that our classification boundary using word-score technique is very close to the actual boundary of the student model. 
This makes our word-score based model a good proxy for the student model and allows us to efficiently craft adversarial samples.

\paragraph{Multiple Tasks.}
We measure the accuracy of our word-score based heuristic for the multi-class dataset of 20Newsgroup. The 20Newsgroup dataset is divided into 6 broader categories and 20 finer categories \cite{20newsdataset}. For example, the `rec.motorcycle' class is a part of the broader class `Recreation'.
Table \ref{tab:20news_6} shows the first five rows for the broader categories and bottom five for the finer classes. To test the word-score based prediction accuracy for each of these classes separately, we select $200$ samples out of which $100$ belong to that category and the remaining $100$ from other categories. Then, the true positive (TP) value for a class refers to the samples belonging to that class which were correctly classified using our word-score mechanism. Similarly, true negative (TN) refers to the correctly classified samples that do not belong to the class. The number of true positives and true negatives are high, indicating that our word-score based heurisitc is effective in the multi-class setting as well. The broader classes have higher accuracy as compared to the finer classes. This observation implies that it becomes difficult to perform adversarial attacks as the number of classes increase (as shown in Section~\ref{sec:wembedding_res}).

\begin{table}
\centering

 \begin{tabular}{c | c c c c } 
 \hline
    & $R_w$ & TP    & TN  & Avg. Acc. (\%)\\ 
 \hline
\glove$_{\tt FT}$ &  1.1 & 1967  & 1931 & 97.4\\
\glove$_{\tt FE}$ & 0.99 & 1698  & 1735  & 85.82\\
 \hline
\end{tabular}
\caption{Accuracy of fake news detection using word-score hypothesis on $2000$ real and $2000$ fake news samples. TP
shows the correctly classified fake news while TN gives the result for correct classification of real news.}
\label{tabel:transfer_learning}
\end{table}

\begin{table}[t]
\centering
\begin{tabular}{c|c c c }
\hline
Categories &  TP  & TN & Avg. Acc. (\%) \\ \hline 
Computer   & 88                       & 100     &       94       \\ 
Recreation & 100                        & 100   &        100            \\ 
Science    & 100                     & 98          &       99 \\ 
Politics   & 100                     & 100         &           100   \\ 
Religion   & 92                         & 100      &         96 \\ \hline  
rec.motorcycles        & 77                       & 92     &   77.46                 \\ 
rec.sport.baseball     & 96                       & 98     &     96.49             \\
talk.politics.guns     & 52                       & 100     &      52.5          \\ 
talk.politics.misc     & 48                        & 97      &       48.4         \\ 
soc.religion.christian & 91                        & 88    &         91.44           \\ \hline      
\end{tabular}
\caption{Accuracy of word-score based prediction for the Newsgroup dataset for broader (top 5) and finer classes (bottom 5) classification tasks. 
Each class is evaluated for true positives and true negatives out of 200 samples (100 of the same class and 100 from other classes).
}
\label{tab:20news_6}
		\vspace{-0.5cm}
\end{table}

\subsection{Shadow Word-score Model}
To use our word-score based prediction mechanism, we have to calculate the word-score for each word in the vocabulary. This ideally requires a large number of queries proportional to the vocabulary size (typically 50,000) to the student model. To reduce the queries to the student model, we build a shadow model $S'$ that tries to mimic the original student model $S$. The initial layer of the shadow model $S'$ is the same as the teacher model $T$ of the student model $S$.
Our threat model assumes the attacker knows the type of the teacher model used in the student model. The objective of the model $S'$ is to learn the weights after the initial layers initialized from $T$. Therefore, we query the model $S$ for a few words and compute the score for these words as per Eq.~\ref{eqn:einstein} above. We use these words and their score as the input dataset to train our shadow model $S'$ to predict the word score for rest of the words in the vocabulary. It should be noted that since word's distribution follow the Zipf's law, the most common words constitute the majority of the usage. 
Hence, we query the word-score for the most common words from $S$ model and use that to train the model $S'$. This strategy allows us to compute the exact word-score for most of the words that may occur in the input text and estimate the score for others using $S'$. 

\paragraph{Evaluating the Shadow Model.} We aim to evaluate the accuracy of our shadow model when trained using different number of word-score pairs for the Fake News Detection student model.
 The total number of words in the vocabulary of the student model are $43,348$.  We query all the words to the student model to get the ground truth values.  The accuracy of the shadow model is calculated over all the words except those used to train the model.  Table~\ref{tabel:shadow} shows the accuracy of our shadow model trained using $10$, $100$, $1000$ and $10,000$ word-score pairs.  We highlight that a limited set of queries ($100$ or $1000$) to the student model are sufficient to
 train a reasonable shadow word-score model. Moreover, these queries are required only once to establish the classification boundary and no queries are required while performing the actual online attack. Our shadow model technique
 distinguishes our attacks from prior work that require unlimited queries to the victim model~\cite{alzantot2018generating,li2018textbugger,lei2019discrete}.
  As shown in Table \ref{tabel:shadow}, the shadow model has the lowest accuracy when trained using scores from a randomly initialized student model. This is expected as this model gains no information from the teacher model. The \glove$_{\tt FT}$ model has lower accuracies as compared to when the teacher model is purely used as a feature extractor \glove$_{\tt FE}$. This is because the shadow model does not have access to the update embedding vector in the case of the \glove$_{\tt FT}$ model. Note that, throughout the process of training a shadow word-score model, we do not require access to the training dataset.

\begin{table}[t]
\centering
 \begin{tabular}{c | c c c  c} 
 \hline
   Model / Queries & 10 & 100 & 1000 & 10000 \\ 
 \hline
 Random & 51.9 & 52.2 & 52.5 & 56.6 \\
\glove$_{\tt FT}$ & 59.9 & 61.5 & 62.4 & 73.2 \\ 
\glove$_{\tt FE}$  & 67.1 & 78.2 & 89.3 &  95.5 \\
 \hline
\end{tabular}
\caption{Table comparing the accuracy of shadow model against all words in the vocabulary of the student model for Fake News Detection. The increase in queries implies more accuracy. Also, \glove$_{\tt FE}$ performs better  than \glove$_{\tt FT}$ as the fine-tuned weights are inaccessible to the shadow model}
\label{tabel:shadow}
\end{table}

\subsection{Creating Adversarial Samples}

Our goal is to generate an adversarial example $x'$ such that it is similar to the actual input $x$ but is classified differently from $x$. Without loss of generality, let us assume that $x'$ should get classified to class $B$. To generate such an example, we use the word-score based prediction as discussed earlier. The word-score based method  states that higher the sum of word scores of class $B$, more likely the model will classify the input as $B$. Therefore, we can search for $x'$ such that $T(x')$ is close to $T(x)$ but has a higher class $B$  score. 

%



%

\begin{algorithm}[t]
\footnotesize
\caption{Adversarial example generation for sentence s containing $w_1 \ldots w_n$ from class A to B }\label{advex_glove}

  \begin{algorithmic}[1]
  \State $S_m$ $\gets$ Shadow model trained using q queries
  \State
  \Function{CheckConstraints}{$w, w'$}
     \State POSw $\gets$ Part of speech of w
   \State POSw' $\gets$ Part of speech of w'
   \If{ POSw = Adjective or POSw = Noun or POSw = Verb or POSw = Aderb}
             \If{POSw = POSw'}
            \State\Return true
            \EndIf
        \EndIf
    \State\Return false
    \EndFunction
    \State
    
   \Function{getReplacementWord}{$w$}
    \State candidateWords $ \gets$ top $g_w$ closest \glove{} words
       \For{\texttt{w' in candidateWords}}
       \If {CheckConstraints($w,w'$) = false}
        \State remove w' from candidateWords
        \EndIf
       \EndFor
	   \If{candidateWords = None}
            \State\Return None
        \EndIf   
        \State $w_r$ $\gets$ candidateWord with max $K_B$ based on $S_m$
        \If{$K_B[w]>K_B[w_r]$}
        \State \Return None
       \EndIf
        \State \Return $w_r$
        
   \EndFunction
   \State
    \Function{generateAdvExample}{$s,g_w,th$}
   \State $t = $(len(replaced words)) / (len(input words))
   \State targetWords $\gets$ Sort $w_i$ based on $K_{A}[w_i]$ from $S_m$
    \For{\texttt{w in targetWords}}
      \If{ $t>th$} 
      \State break
	\EndIf
	\State $w_r$ $\gets$ getReplacementWord($w$) 
	\If  {$w_r != None$} 
        \State Replace w with $w_r$
         \State Update t
	\EndIf    
    \EndFor
   \EndFunction
  \end{algorithmic}

\end{algorithm}
\setlength{\textfloatsep}{4pt}


The attack in Algorithm \ref{advex_glove} considers the \glove{} word embedding as the teacher model to attack a text classification student model initialized by \glove{} vectors. A shadow model is initially trained using $q$ queries to the student model and observing the output probabilities. We use the trained shadow model thereafter, to retrieve word score for all the required words. The adversarial example generation algorithm (function {\tt generateAdvExample}) takes two parameters along with the original input example. First parameter is the number of similar words considered for the attack based on the \glove{} model, $g_w$. The second parameter is defined as the threshold $th$ or the maximum fraction of words allowed to be replaced by similar words when performing the adversarial attack. When choosing a replacement word for an input word as shown in function {\tt getReplacementWord}, we chose the word to be among $g_w$ closest word according to the \glove{} vectors with maximum class B word score. Selecting nearby words in the vector embedding space guarantees semantic correctness of the generate adversarial example.
 While selecting the replacement word, we satisfy the constraints described in function {\tt checkConstraints}. The constraint procedure makes sure that the replacement word is among the following part-of-speech: noun, adjective, verb or adverb and has the same part-of-speech as the input word. This guarantees that replacing the input word with similar words does not change the structure of the input and hence retains its syntactic similarity.  The {\tt generateAdvExample} function sorts the words in the input sentence based on the score of Class A. Replacing high scoring words from class A reduces the sum score of class A significantly. If the fraction of replaced words $t$ exceed the specified threshold $th$ or there are no more nearby words i.e., {\tt targetWords}, then the algorithm terminates.

 \section{Word Embedding attack evaluation}
 \label{sec:wembedding_res}

We present a detailed evaluation of our adversarial examples generated using attack Algorithm~\ref{advex_glove} against different types of downstream tasks. We evaluate the attack accuracy of our algorithm against the best-known prior works. We also analyze our algorithm's attack accuracy in misclassifying inputs from a source class to a target class in a multi-class scenario.

\subsection{Experimental Setup}
\paragraph{Pre-trained model  \& Teacher model.}
We consider the Global Vectors or \glove{} model as our teacher model which is among the best word representation models\cite{pennington2014glove}. 
\glove{} model is available online and is trained on a corpus of Wikipedia data with 6B tokens, 400K vocab, uncased, 50d, 100d, 200d, \& 300d vectors.
We use it with 300-dimensional vector in our experiments.

\paragraph{Dataset  \& Downstream tasks.}
We consider three different downstream tasks that incorporate both binary and multi-class classification student models
 trained using \glove{} teacher model. The student model for each of these tasks is a $1$-layer LSTM with $256$ hidden units followed by a linear layer trained using a learning rate of $0.001$.  The three tasks are as follows:
 
\begin{itemize}
\squish
\item {\em Fake News Detection~\cite{fakenewsdataset}.} This is a binary classification task to classify whether news articles are real or fake.  The dataset has of 5335 training and 1000 test news reports.

\item {\em IMDB Movie Reviews~\cite{maas-EtAl:2011:ACL-HLT2011}.} The IMBD dataset contains movie reviews with sentiment polarity labels. A review is negative if it has   a score of less than 5 out of 10, while a positive review has a score greater than 6. IMBD dataset consists of 25000 training and test movie reviews.

\item {\em 20 Newsgroup~\cite{20newsdataset}.}   The 20Newsgroup dataset consists of 20 classes of newsgroups  ranging from sports to politics to religion. This dataset also has a broader classification of documents into 6 groups. The training and test dataset consists of 11270 and 7503 samples respectively.
\end{itemize}


%
%

\subsection{Student Model Accuracy}

\begin{table}[t]
\centering
\begin{tabular}{c | c}
\hline
Model                    & Accuracy        \\ \hline
\multicolumn{2}{c}{Fake News Detection}    \\ \hline
\glove$_{\tt FT}$         & 95.4            \\ 
\glove$_{\tt FE}$         & 95.1            \\ 
\glove$_{\tt DR}$         & 95.8            \\ 
BERT$_{\tt FT}$                     & 92.1            \\ \hline 
\multicolumn{2}{c}{IMDB Movie Reviews}         \\ \hline
\glove$_{\tt FT}$         &     90.3            \\ 
\glove$_{\tt Dr}$         &     90.3        \\ \hline  
\multicolumn{2}{c}{20Newsgroup} \\ \hline
\glove$_{\tt FT}$ - 6 class          & 92.1            \\ 
\glove$_{\tt FT}$- 20 class         &  67.3           \\ \hline
\end{tabular}%
\caption{Model accuracy for different model's trained on different datasets. \glove$_{\tt FT}$ is a fine-tuned student model , \glove$_{\tt FE}$ is a feature extractor model. \glove$_{\tt DR}$ is same as \glove$_{\tt FT}$ with dropouts in between layers.} 

\label{tab:model_acc}
\end{table}

We train three variants of the student model
with \glove{} as a feature-extractor called \glove$_{\tt FE}$, with fine-tuning called \glove$_{\tt FT}$ and fine-tune with dropouts  \glove$_{\tt DR}$.  We report the original model accuracy for the student models in Table \ref{tab:model_acc}\footnote{Since \glove$_{\tt FT}$ performs better than \glove$_{\tt FE}$, we use \glove$_{\tt FE}$ in IMDB and 20newsgroup dataset. }. The Fake News Detection accuracy is the highest for \glove$_{\tt DR}$ model as adding dropouts to the model makes the model more generalization and achieves better accuracy on the test dataset. BERT model is initialized by BERT pre-trained vector followed by a simple feed-forward layer. We discuss the architecture in more details in Section \ref{sec:sembedding}. 
We train the IMBD dataset with \glove$_{\tt FT}$ and \glove$_{\tt Dr}$, to compare the results with prior best attacks, as discussed later in the section. The model accuracy is the same with and without dropouts because the IMDB dataset has a larger training dataset that prevents overfitting. We also explore the multi-class setting using the 20Newsgroup dataset. The model trained on this dataset has a similar configuration as \glove$_{\tt FT}$, except that the final output layer was equal to the number of classes. We report accuracy for 6 classes and 20 classes in the table. The model accuracy decreases with an increase in the number of classes, as the 20 classes are the finer categorization of the 6 classes. Although we did not optimize our models to the  state-of-the-art classification accuracy they are close to the best-known prediction numbers.

\subsection{Attack Accuracy for Binary Tasks}
\begin{table}[t]
\centering
\begin{tabular}{l|c c|c c}
\hline

\multicolumn{1}{c|}{$g_w$} & \multicolumn{2}{c|}{Fake News Detection}                                & \multicolumn{2}{c}{IMDB  Movie Reviews}                                                     \\ \hline
                            & \multicolumn{1}{c}{Real to Fake} & \multicolumn{1}{c|}{Fake to Real} & \multicolumn{1}{c}{+ve. to -ve} & \multicolumn{1}{c}{-ve. to +ve} \\ \hline
\multicolumn{5}{c}{Model trained without Dropout} \\
\hline

10                          & 83.5                              & 56.2                              & 85.8                                      & 82.8                                      \\ 
20                          & 89.3                              & 70.5                              &   94.2                                       &  95.9                                        \\ \hline
\multicolumn{5}{c}{Model trained with Dropout} \\
\hline
10                          & 72.6                           & 60.9      &  89.6&        83.9  \\ 
20                          & 75.7                          & 78.1     &   95.9   &    97.5  \\ \hline

\end{tabular}
%
\caption{Attack accuracy with a perturbation threshold $th$ of 0.2.}
\label{tab:attack_binary}
\end{table}

\begin{table}[t]
\centering
 \begin{tabular}{c |c| c |c  c} 
 \hline
    Model & $g_w$ & th &  \multicolumn{2}{c}{Attack Accuracy}\\ 
    \cline{4-5}
    & & & Real to Fake & Fake to Real \\
 \hline 
 \multirow{ 6}{*}{\glove$_{\tt FE}$} & \multirow{ 3}{*}{10}& 0.2 & 62.7  & 48.3 \\ 
 
 	&  & 0.5 & 81.1 & 63.4 \\
 	 & & 1 & 81.1 & 64.4 \\
 \cline{2-5}
 	& \multirow{ 3}{*}{20} & 0.2 & 75.3 & 54.1 \\
 	&  & 0.5 & 92.2 & 79.6 \\
   &  & 1 & 92.2  & 80.6 \\
 \hline
\multirow{ 6}{*}{\glove$_{\tt FT}$  } & \multirow{ 3}{*}{10}& 0.2& 83.5  & 56.2\\ 
 	&  & 0.5 & 88.8 & 80.2 \\
 	& & 1 & 88.8 & 81.2\\
 \cline{2-5}
 	& \multirow{ 3}{*}{20} & 0.2 & 89.3 & 70.5 \\
 	&  & 0.5 & 96.2 &  90.3\\

   &  & 1 &  96.8 & 91.3 \\
 \hline
   
\end{tabular}
\caption{Table comparing the accuracy of adversarial attack against different models. $g_w$ refers to the number of closest Glove words considered for replacement. $th$ refers to upper limit on fraction of words replaced.
}
\label{tabel:acc}
\end{table}


\begin{table}[t]
\centering
\begin{tabular}{c| c c c c c }
\hline
$g_w$\textbackslash{}Classes & Comp & Rec   & Sci  & Religion  & Misc  \\ \hline
100                          & 0 & 99  & 68 & 51 & 36 \\ 
200                          & 1 & 100 & 94 & 87 & 58 \\ 
300                          & 3 & 100 & 97 & 93 & 80 \\ \hline
$Avg.t $ & 0.09 & 0.37   & 0.31  & 0.26  & 0.31  \\ 
\end{tabular}
\caption{Attack accuracy on the Newsgroup for 6 classes dataset. The target class is Politics.  The last row represents average fraction of replaced words across different classes.}
\label{tab:attack_news20_6}
\end{table}

In Table \ref{tab:attack_binary}, we present the attack accuracy on the Fake News Detection and IMDB Movie Reviews student model trained using \glove{} model with and without dropouts. We exclude using \glove{} as a feature extractor since it achieves the lowest validation accuracy. For generating the advesarial examples for these tasks, we use a shadow model trained with $1000$ word-score pairs queried from the student model. As discussed in Section \ref{sec:wembedding}, $g_w$ refers to the number of closest words based on \glove{}, that are considered for replacing the input word. Therefore, $g_w$ is an indicator of the semantic closeness of the adversarial examples to the original input. 
For the Movie Reviews, we consider misclassification from both positive to negative and negative to positive sentiments. We randomly sampled 1000 positive and negative reviews from IMDB test dataset and tried to misclassify them by generating adversarial examples using our attack algorithm. The average attack accuracy with $g_w=20$ for the movie reviews  is $95\%$ and $96.7\%$ without and with dropouts respectively. For the Fake News Detection task, we consider 1000 real and 1000 fake samples and try to misclassify them to the opposite class. The average attack accuracy for the news dataset is $80\%$ and $76.9\%$ without and with dropout respectively. We find that it is easy to misclassify a real news to fake rather than fake to real. This is because the number of fake words, as well as the sum of fake word scores, is more than that of real words. With more alternatives in hand, it is comparatively easier to change real news reports to fake than the other way around. 
We keep the perturbation threshold to be $0.2$ i.e, at max $20\%$ of the words could be replaced to generate the
adversarial example. The attack accuracy is a slightly higher in the case of movie reviews. This might be because of a clear separation between words used to express positive and negative sentiments, compared to fake and real news.
However, we do observe that the word ``Obama'' has a high score for real news and ``Wikileaks'' for the fake news.

We perform a fine grained attack evaluation for the Fake News Detection task by varying the threshold $th$ parameter.
The threshold $th$ is the upper bound on the fraction of words that are allowed to be replaced. 
Tabel \ref{tabel:acc} shows the accuracy for misclassifying both real and fake news with $th$ values of $0.2$, $0.5$ and $1$. 
As expected, an increase in the value of $th$ and $g_w$ indicates an increase in attack accuracy. However, the attack accuracy gets saturated when there are not enough candidate words due to stringent semantic $g_w$ and replacement $th$ threshold. In most cases, increasing the threshold value from 0.5 to 1, doesn't increase the accuracy as there are not enough semantically similar words from \glove{} that satisfy the constraint mentioned in Algorithm \ref{advex_glove}. 

\begin{table}[t]
\centering
\begin{tabular}{ c | c| c l c l c }
\hline
\multicolumn{2}{c}{Category}                                                                    & \multicolumn{2}{c}{$g_w=$100 } & \multicolumn{2}{c}{$g_w=$200 } & Avg. t \\ \hline
\multirow{4}{*}{Recreation} & rec.sport.hockey                                                    & \multicolumn{2}{c}{22}                                                      & \multicolumn{2}{c}{84}                                                     & 0.03 \\ 
                            & rec.motorcycles                                                     & \multicolumn{2}{c}{0 }                                                      & \multicolumn{2}{c}{0 }                                                     &0.004 \\ 
                            & rec.autos                                                           & \multicolumn{2}{c}{6 }                                                      & \multicolumn{2}{c}{6 }                                                      &0.006\\ 
                            & rec.sport.                                                          & \multicolumn{2}{c}{44}                                                      & \multicolumn{2}{c}{94 }                                                      &0.05\\ \hline
\multirow{3}{*}{Religion}   & soc.religion.christian                                              & \multicolumn{2}{c}{39}                                                      & \multicolumn{2}{c}{90}                                                      &0.06\\ 
                            & alt.atheism                                                         & \multicolumn{2}{c}{0 }                                                       & \multicolumn{2}{c}{4}                                                       &0.05\\ 
                            & talk.religion.misc                                                  & \multicolumn{2}{c}{2}                                                       & \multicolumn{2}{c}{4}                                                      &0.02\\ \hline
\multirow{4}{*}{Science}    & sci.crypt                                                           & \multicolumn{2}{c}{18 }                                                      & \multicolumn{2}{c}{24}                                                      &0.02\\ 
                            & sci.space                                                           & \multicolumn{2}{c}{14 }                                                      & \multicolumn{2}{c}{42 }                                                      &0.03\\ 
                            & sci.med                                                             & \multicolumn{2}{c}{92 }                                                      & \multicolumn{2}{c}{96 }                                                      & 0.06\\ 
                            & sci.electronics                                                     & \multicolumn{2}{c}{26 }                                                      & \multicolumn{2}{c}{30 }                                                      &0.04\\ \hline
\multirow{5}{*}{Computer}   & \begin{tabular}[c]{@{}c@{}}comp.os.ms-\\ windows.misc\end{tabular}  & \multicolumn{2}{c}{0}                                                      & \multicolumn{2}{c}{0}                                                      &0.01 \\ 
                            & comp.windows.x                                                      & \multicolumn{2}{c}{0}                                                       & \multicolumn{2}{c}{2}                                                       &0.03\\ 
                            & \begin{tabular}[c]{@{}c@{}}comp.sys.ibm.\\ pc.hardware\end{tabular} & \multicolumn{2}{c}{0}                                                      & \multicolumn{2}{c}{0}                                                      &0.005\\ 
                            & comp.graphics                                                       & \multicolumn{2}{c}{0}                                                       & \multicolumn{2}{c}{16}                                                      &0.04\\ 
                            & \begin{tabular}[c]{@{}c@{}}comp.sys.\\ mac.hardware\end{tabular}    & \multicolumn{2}{c}{0}                                                          & \multicolumn{2}{c}{0}                                                          &0\\ \hline
Misc                        & misc.forsale                                                        & \multicolumn{2}{c}{74}                                                      & \multicolumn{2}{c}{96}                                                      &0.06\\ \hline
\multirow{2}{*}{Politics}   & \begin{tabular}[c]{@{}c@{}}talk.politics.\\ midwest\end{tabular}    & \multicolumn{2}{c}{98 }                                                      & \multicolumn{2}{c}{100 }                                                     & 0.04\\ 
                            & talk.politics.guns                                                  & \multicolumn{2}{c}{16}                                                      & \multicolumn{2}{c}{84}                                                      &0.03\\ \hline
\end{tabular}%
\caption{Attack accuracy in converting `talk.politics.misc' to different classes. Avg. t is the fraction of replaced words in the input for each category.}
\label{tab:attack_20news}
\end{table}

\subsection{Multi-class Classification}

 We analyze the effect of increasing classes on the efficacy of our attack algorithm using the word-score based prediction.
Prior works have not shown adversarial examples for multiple classification task with the exception of Liang et al. \cite{liang2017deep}. However, they do not perform a comprehensive study and limit their evaluation to only 23 selected examples from different classes and change it to a randomly selected class. In contrast, we perform an exhaustive evaluation using the 20Newsgroup dataset. To understand the impact of increase in the output classes, we divide the dataset into 6 and 20 classes.  The 6 classes consist of `Computer', `Recreation', `Science', `Miscellaneous', `Politics' and `Religion'.  The 20 classes comprise of finer divisions of each of the 6 classes. For example, `Politics' is divided into `talk.politics.misc', `talk.politics.guns' and `talk.politics.mideast'. All our adversarial examples are generated using a shadow model trained with only $1000$ queries to the student model.

\begin{table*}[t]
\centering
\begin{tabular}{c|c c c c c }
\hline
Method                   & Threat Model    & Queries             & Targeted/untargeted & Model         & Success rate \\ \hline
\textbf{Word-Score (This paper)} & \textbf{Grey-box} & \textbf{Restricted} &  \textbf{Targeted} &  LSTM  &  {\bf 96.7\%}    \\ 
Alzantot et al.\cite{alzantot2018generating}          & Black          & Unlimited           & Targeted            & LSTM          & 97\%         \\ 
iAdv-Text \cite{sato2018interpretable}              & White          & Unlimited           & Non-targeted        & LSTM          & 93.92\%      \\ 
TextBugger \cite{li2018textbugger}              & White          & Unlimited           & Non-targeted        & char-CNN      & 86.7\%       \\ 
Samanta et al.\cite{samanta2017towards}          & White          & Unlimited           & Non-targeted        & CNN           & 67.45\%      \\ 
Gong et al.\cite{gong2018adversarial}            & White          & Unlimited           & Targeted            & CNN           & 86.66\%      \\ \hline
\end{tabular}%
\caption{Comparing our method, Word-score approach, to the best-known previous attack in literature on IMBD Movie Review dataset.}
\label{tab:comp-attack}
\end{table*}

\begin{table*}[t]
\centering
\resizebox{\textwidth}{!}{%
\begin{tabular}{|c|c|c|}
\hline
\textbf{Model}           &  \begin{tabular}[c]{@{}c@{}}\textbf{Prediction},\\ \textbf{Confidence(-ve,+ve)}\end{tabular} & \textbf{Text}                                                                                                                                                                                                                                                                                                                                                                                                                \\ \hline \hline
\textbf{Original}        & \textbf{NEG,(0.99,0.01) }                                               & \begin{tabular}[c]{@{}c@{}}``This movie had terrible acting, terrible plot, and terrible choice of actors. \\ (Leslie Nielsen ...come on!!!) the one part I considered slightly funny was the battling FBI/CIA agents, \\ but because the audience was mainly kids they didn’t understand that theme"\end{tabular}                                                                                                   \\ \hline
\textbf{Word-Score(Our)} & \textbf{POS,(0.1,0.9)}                                                  & \begin{tabular}[c]{@{}c@{}}``This movie had \textbf{unspeakable} directing, \textbf{unspeakable} assassination, and \textbf{unspeakable} choice of characters. \\ (Leslie Nielsen ...come on!!!) the one part I considered \textbf{moderately} funny was the battling FBI/CIA agents, \\ but because the audience was mainly kids they didn’t understand that theme"\end{tabular}                                                                        \\ \hline
Alzantot \cite{alzantot2018generating}         & POS,(0.2,0.8)                                                  & \begin{tabular}[c]{@{}c@{}}``This movie had \textbf{horrific} acting, \textbf{horrific} plot, and textbf{horrifying} choice of actors. \\ (Leslie Nielsen ...come on!!!) the one part I \textbf{regarded} slightly funny was the battling FBI/CIA agents, but \\ because the audience was mainly \textbf{youngsters} they didn’t understand that theme."\end{tabular}                                                                                            \\ \hline\hline
\textbf{Original}        & \textbf{NEG,(0.96,0.04) }                                               & \begin{tabular}[c]{@{}c@{}}``I watched this movie recently mainly because I am a Huge fan of Jodie Foster's. I saw this movie was made\\  right between her 2 Oscar award winning performances, so my expectations were fairly high. Unfortunately, \\ I thought the movie was terrible and I'm still left wondering how she was ever persuaded to make this movie. \\ The script is really weak."\end{tabular}      \\ \hline
\textbf{Word-Score(Our)} & \textbf{POS,(0.3,0.7)}                                                  & \begin{tabular}[c]{@{}c@{}}``I watched this movie recently mainly because I am a \textbf{enormous} fan of Jodie Foster's. I saw this movie was \\ made right between her 2 Oscar award winning performances, so my expectations were fairly high. Really, \\ I thought the movie was \textbf{horrible} and I'm still left wondering how she was ever persuaded to make this movie. \\ The \textbf{screenplay} is really \textbf{sluggish}."\end{tabular} \\ \hline
TextBugger \cite{li2018textbugger}     & POS,(0.4,0.6)                                                  & \begin{tabular}[c]{@{}c@{}}``I watched this movie recently mainly because I am a Huge fan of Jodie Foster's. I saw this movie was made \\ right between her 2 Oscar award winning performances, so my expectations were fairly high. \textbf{Unf0rtunately}, \\ I thought the movie was \textbf{terrib1e} and I'm still left wondering how she was ever persuaded to make this movie. \\ The script is really \textbf{wea k}."\end{tabular}     \\ \hline \hline
\textbf{Original}        & \textbf{NEG,(0.97,0.03) }                                                & \begin{tabular}[c]{@{}c@{}}``A sprawling, overambitious, plotless comedy that has no dramatic center. It was probably intended to have an\\  epic vision and a surrealistic flair (at least in some episodes), but the separate stories are never elevated into a \\ meaningful whole, and the laughs are few and far between. Amusing ending, though. (*1/2)"\end{tabular}                                          \\ \hline
\textbf{Word-Score(Our)} & \textbf{POS,(0.1,0.9) }                                                 & \begin{tabular}[c]{@{}c@{}}``A sprawling, overambitious, \textbf{balanchine} comedy that has no \textbf{remarkable} center. It was probably designed to have \\ an epic vision and a \textbf{impressionistic} flair (at least in some episodes), but the separate tales are never elevated into a \\ meaningful whole, and the laughs are \textbf{several} and far between. Amusing ending, though. (*1/2)"\end{tabular}                                \\ \hline
Samanta\cite{samanta2017towards}      & NEG,(0.97,0.03)                                                & \begin{tabular}[c]{@{}c@{}}``A sprawling, overambitious, plotless \textbf{funny} that has no dramatic center. It was probably intended to have an epic \\ vision and a surrealistic fair (at least in some episodes), but the separate stories are never elevated into a \textbf{greatly} meaningful \\ whole, and the laughs are \textbf{little} and far between. Amusing ending, though. (*1/2)"\end{tabular}                                   \\ \hline
TextFool\cite{liang2017deep}       & NEG,(0.99,0.01)                                                & \begin{tabular}[c]{@{}c@{}}``A sprawling, overambitious, plotless \textbf{horrorible} that has no dramatic center. It was probably intended to have an \textbf{fail} \\ vision and a surrealistic \textbf{fair} (at least in some episodes), but the separate stories are never elevated into a \textbf{false} meaningful whole, \\ and the laughs are few and far between. Amusing ending, though. (*1/2)"\end{tabular}                                      \\ \hline
\end{tabular}%
}
\caption{Examples of adversarial samples crafted from IMDB Movie Reviews dataset using our algorithm and some of the prior best works}
\label{tab:comp-adv}
\vspace{-0.5cm}
\end{table*}

 In Table \ref{tab:attack_news20_6}, we show the attack accuracy for misclassifying 100 test documents from Politics to the 5 other classes mentioned in the table.  We achieve an average attack accuracy of $75\%$ for our targeted misclassification attacks.
 We can misclassify documents from politics to recreation with almost $100\%$ accuracy. This is because of the presence of a large number of candidate replacement words. However, the attack accuracy for converting politics to the computer category is the lowest, which is expected as there are not enough similar words for high scoring political words that have a significant score in computer class. The average fraction of replaced words is less than 0.09 for this case. For other categories, the average fraction is around 0.3 which indicates that  fewer words are perturbed to perform the misclassification. Similar to binary classification, we observe that increasing the value of $g_w$, increases the number of potential replacement words, leading to an increase in the attack accuracy. This shows that it is easier to convert to classes that share similar vocabulary, compared to classes with different word distribution.

 Table \ref{tab:attack_20news} shows the attack accuracy for converting `talk.politics.misc' to the other $19$ classes. Similar to Table \ref{tab:attack_news20_6}, the attack accuracy is lowest for classes falling under the computer category. Also, changing to other sub-classes within the politics category has a higher success rate compared to other classes, since they share a common vocabulary. With the increase in the number of classes, the number of potential replacement words decrease resulting in smaller $t$ values (last column). Also, increasing the value of $g_w$ leads to higher attack accuracy similar to the previous table. It is interesting to note that there is a jump in attack accuracy for the Misc group compared to Table \ref{tab:attack_news20_6}, although the accuracy remains similar for other classes. This is because the inputs in `talk.politics.misc' are more similar to the Misc group than the whole Politics category in general. Thus, although  misclassifcation accuracy for some individual classes is high such as $98\%$ for ``talk.politics.midwest'', the average attack accuracy is only $41\%$. 

\subsection{Comparison to Previous Work}

 In Table \ref{tab:comp-attack}, we compare our attack accuracy to best-known attacks for generating
adversarial examples for non-tranfer learned Movie Reviews classifier, as reported in a recent survey paper by Wang et al. \cite{wang2019survey}.  The {\tt threat model} column in Table \ref{tab:comp-attack} captures the difference in the assumption about the adversarial setting. Our attack is very close to the best-known attack accuracy by Alzantot et al. \cite{alzantot2018generating}. Alzantot et al. consider a black-box setting but assume unlimited queries to the black-box model.
We highlight that ours is the only attack that works with limited queries to the victim model because of the grey-box threat model 
that is enabled with transfer learning. Our attack algorithm is efficient with only 7 sec required for generating an example as compared to 43 sec by Alzantot et al~\cite{alzantot2018generating}.  This is an artifact of our systematic approach of generating adversarial examples on text as compared to randomly trying different words. Our attack outperforms all other prior work in the white-box setting. This indicates that the transfer learning setting indeed increases the susceptibility towards adversarial attacks for text-based classification models.

Finally, we take adversarial examples from some of these prior works and show how they compare against our examples in Table \ref{tab:comp-adv}. We report the confidence value of our student model in column 2. The adversarial examples by TextBugger use spelling error and spaces between words, which can be detected by spell check\cite{li2018textbugger}. Adversarial examples by Samanta et al. \cite{samanta2017towards} and TextFool \cite{liang2017deep} did not get misclassified on our model. Lastly, the example by Alzantot et al. \cite{alzantot2018generating} is similar to the adversarial example generated by our attack.

\section{Sentence embedding attacks}
\label{sec:sembedding}
In this section, we explore attacks on student models trained using sentence-level embedding models. These models represent the input sentence into an n-dimensional vector that captures the context of the entire input. For generating adversarial examples on these models, we take into account some other unintended features that the model learns, for e.g., the length of the input or how it memorizes sentences. We initially show how our hypothesis is correct by showing the presence of these biases in the training dataset or in the teacher model and then describe the attack algorithm. We show the efficacy of these attacks on state-of-the-art BERT model implying that even these models are not robust against misclassification attacks.  

\paragraph{Teacher Model: Bert-Base.}
BERT, stands for Bidirectional Encoder Representations from Transformers, which is a sentence level embedding vector representation developed by Google. BERT achieves state-of-the-art performance on a large number of sentence and word level tasks, outperforming previous task-specific architectures. Unlike word-level embedding model which are context-free, BERT is a bi-directional contextual representation. That means, while representing a word $w_2$ in a sentence $w_1 w_2 w_3$, BERT not only take the previous words $w_1$ into consideration but also $w_3$. BERT models have tens of layers and millions of parameters. The BERT model took 4 days to train on 4 to 16 Cloud TPUs \cite{devlin2018bert}. Fine-tuning of BERT model is relatively cheaper and takes a few hours to train on GPU. We use the BERT model, ``bert-base-uncased"  that has 12-layer, 768-hidden, 12-heads, 110M parameters. We train our student model for classification of real-fake news on top of this pre-trained model. 
 
 \paragraph{Student model.}
 We train the Fake News Detection binary task using the \bert$_{\tt FT}$ as the teacher model.
 The student model has the initial layers of ``bert-base-uncased" model, which is connected to a  simple feed-forward layer that outputs the prediction.  We fine-tune the \bert{} model with a learning rate of 0.000001 and a learning rate of 0.001 for the final feed-forward layer. The student model accuracy is $92.1\%$.
 
 \paragraph{Identifying unintendedn features.} 
 We observe that the BERT pre-trained model is robust against the word-score based attack. This is expected as the BERT model takes the whole sentence as input and output vectors representations based on its context. However, the model is susceptible to other unintended features.  We identify two unintended features, length, and sentence-based features, in the model and use it to generate adversarial examples. These two methods are discussed in detail in the following subsections. 

\subsection{Length-based Adversarial Examples}
The length-based adversarial examples take advantage of the fact that the model is making predictions based on an unintended feature - the length of the input. The BERT pre-trained captures the length of the sentence in the vector representation, which is then used as a feature in the classification task. However, the length of the input is invariant in the task of real and fake news classification. A fake news report can consist of any number of sentences. However, we will show that this feature is present in the training input distribution.

 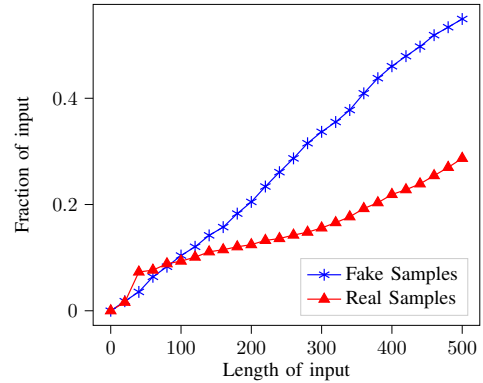
\begin{figure}[t]
 \centering
 \begin{tikzpicture}[scale=0.75]

\begin{axis}[
legend cell align={left},
legend style={at={(0.97,0.03)}, anchor=south east, draw=white!80.0!black},
tick align=outside,
tick pos=left,
x grid style={white!69.01960784313725!black},
xlabel={Length of input},
xmin=-25, xmax=525,
xtick style={color=black},
y grid style={white!69.01960784313725!black},
ylabel={Fraction of input},
ymin=-0.0274612693653173, ymax=0.576686656671664,
ytick style={color=black}
]
\addplot [semithick, blue, mark=asterisk, mark size=3, mark options={solid}]
table {%
0 0
20 0.0179910044977511
40 0.0354822588705647
60 0.063968015992004
80 0.0829585207396302
100 0.103948025987006
120 0.120439780109945
140 0.141929035482259
160 0.157421289355322
180 0.182908545727136
200 0.204897551224388
220 0.233883058470765
240 0.260869565217391
260 0.286856571714143
280 0.315342328835582
300 0.336831584207896
320 0.355322338830585
340 0.377811094452774
360 0.409295352323838
380 0.437781109445277
400 0.460269865067466
420 0.479260369815092
440 0.497251374312844
460 0.518740629685157
480 0.533733133433283
500 0.549225387306347
};
\addlegendentry{Fake Samples}
\addplot [semithick, red, mark=triangle*, mark size=3, mark options={solid}]
table {%
0 0
20 0.015992003998001
40 0.0729635182408796
60 0.0764617691154423
80 0.088455772113943
100 0.0934532733633183
120 0.100949525237381
140 0.110944527736132
160 0.114942528735632
180 0.120439780109945
200 0.124437781109445
220 0.132433783108446
240 0.135932033983008
260 0.142428785607196
280 0.147926036981509
300 0.15592203898051
320 0.16591704147926
340 0.176911544227886
360 0.19240379810095
380 0.203398300849575
400 0.218890554722639
420 0.227886056971514
440 0.23888055972014
460 0.254372813593203
480 0.269865067466267
500 0.286856571714143
};
\addlegendentry{Real Samples}
\end{axis}

\end{tikzpicture}
\caption{ Comparing fraction of training real and fake samples having length less than x.   }
\label{fig:sent_len}
\end{figure}
 
 \paragraph{Evaluation Methodology.}
 In Figure \ref{fig:sent_len}, we show that the number of fake samples in the training dataset with length less than 512 is almost double of that of real samples. We consider the length limit of 512, as BERT model can take a maximum input length of 512.
 Of all the training samples, $72\%$ of real samples had a length of 512, while only $45\%$ of fake samples had a length of 512. This introduces an unintended useful feature, which is the length of the input. The student model uses this feature to classify news reports into fake and real. We can, therefore, generate adversarial examples by reducing the length of real news articles to misclassify them as fake.

\begin{figure}[t]
\centering
\begin{tikzpicture}[scale=0.75]

\definecolor{color0}{rgb}{0.75,0.75,0}

\begin{axis}[
legend cell align={left},
legend style={at={(0.97,0.03)}, anchor=south east, draw=white!80.0!black},
tick align=outside,
tick pos=left,
x grid style={white!69.01960784313725!black},
xlabel={Number of sentences},
xmin=-0.3, xmax=6.3,
xtick style={color=black},
xtick={0,1,2,3,4,5,6},
xtick={0,1,2,3,4,5,6},
xticklabels={1,2,3,4,5,6,$>$6},
xticklabels={1,2,3,4,5,6,$>$6},
y grid style={white!69.01960784313725!black},
ylabel={Accuracy},
ymin=7.6, ymax=104.4,
ytick style={color=black}
]
\addplot [semithick, color0, mark=triangle*, mark size=3, mark options={solid}]
table {%
0 12
1 17
2 28
3 39
4 43
5 64
6 89
};
\addlegendentry{Real Samples}
\addplot [semithick, blue, mark=*, mark size=3, mark options={solid}]
table {%
0 100
1 100
2 99
3 99
4 99
5 99
6 90
};
\addlegendentry{Fake samples}
\end{axis}

\end{tikzpicture}
\caption{ Comparing the change in model accuracy with different input length. We show that decreasing the number of sentences in the input increases the probability of being classified as fake for both real and fake samples.}
\label{fig:length_accuracy}
\end{figure}
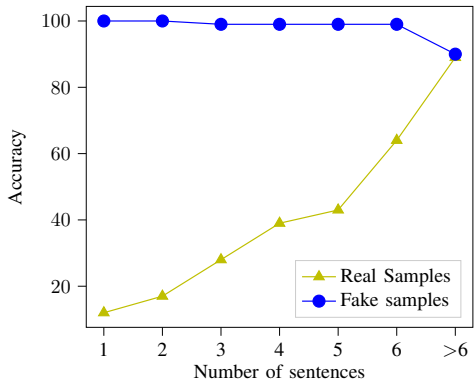

\paragraph{Attack Results.}
We develop a simple adversarial attack to exploit the feature that the model is making classifications based on the length of the input. We take 100 real and fake samples and reduce the number of input sentences. Reducing the number of sentences in a report should not affect the classification of the input. In Figure \ref{fig:length_accuracy}, attack accuracy for unmodified real and fake samples is around $90\%$, which is also close to the model accuracy. However, as we decrease the length of the input, the model accuracy corresponding to real samples reduces to $12\%$. On the other hand, the accuracy of fake samples increases to $100\%$ by decreasing the number of sentences. This is expected as the model learns from the training set distribution that smaller length input has more probability of being fake news than real news.

\subsection{Sentenced-based Adversarial Examples}

This adversarial attack is based on the model memorizing sentences from the inputs. The BERT model outputs an n-dimensional vector representation for a sentence, which is the input to the linear classification layer. Thus, the main classification is based on the output of the BERT model. During the training process, we retrain the BERT layer with very low training rate (0.000001) and therefore even though an adversary does not have access to the BERT layer in student model, the output of the BERT layer in the student model should be very similar to the output of the pre-trained BERT model. As the classification is based on the output of BERT, adding sentences that make the vector representation of BERT closer to the vector representation of some real sample, should also change the prediction of the student model towards real. 

\paragraph{Method.} We take a real and a fake news report that has been classified by the student model correctly with high confidence. Let the vector representation from the BERT model of the real report be $v_r$ and of the fake news report be $v_f$. The goal is to generate fake samples that are closer to $v_r$ and vice-versa. Since the student model classification is dependent on the BERT representation, the student model will then misclassify a fake sample as real. To show that this hypothesis is indeed true, we take 100 real samples and 100 fake samples and calculate their dot product with $v_r$ and $v_f$. $76\%$  real samples were closer to $v_r$ compared to $v_f$ and $78\%$  fake samples were closer to $v_f$ compared to $v_r$. Thus, we  generate adversarial samples that have fake content but vector representation close to $v_r$. 

\begin{figure}[t]
\centering
\includegraphics[width=0.9\linewidth]{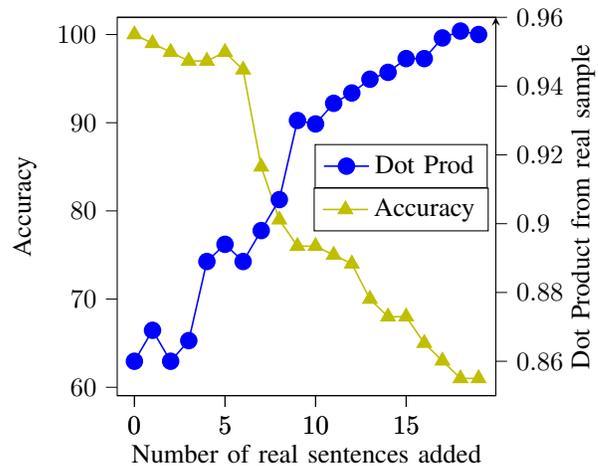}
\caption{ Comparing the change in model accuracy with different input length. This graph shows that moving the BERT representation of the fake samples closer towards the BERT representation of a known real sample, increases the chance of get classified as real.  }
\label{fig:sent_accuracy}
\end{figure}

\paragraph{Results.}
The adversarial attack setting is to misclassify 100 fake news samples to real by trying to move the dot product closer to that of a real sample. We achieve this by adding one sentence at a time from the real sample to each of the test fake samples. Since the maximum length of inputs is 512, we consider fake samples that have a length of less than 200. This ensures that the maximum length does not exceed 512. As seen in Figure \ref{fig:sent_accuracy}, adding sentences from a real sample moves the average dot product closer to that of the real sample. Simultaneously, the accuracy of the model decreases from 100 to 61. The slope of the line representing the average dot product of the samples and the accuracy are complementary to each other. As the dot product increases significantly between 5 and 10, so does the accuracy drops massively.

 \section{Defenses}

We explore the effectiveness of our attacks on some of the potential defenses.  We select three of the common defenses from the literature against adversarial attack, which applies to our transfer-learning setting.  

\subsection{Fine-tuning all layers}
 Recently, transfer learned models have been shown to be susceptible to different adversarial and poisoning in images~ \cite{shafahi2018poison}, \cite{yao2019latent}. Their transfer learning approach assumes that initial $k$ layers are kept frozen and used purely as feature extractors. One of the obvious defense against attacks exploiting transfer-learned features is to retrain the initial layers with the goal to erase the memorized features in the teacher model. However, this has shown to decrease the accuracy since the high capacity memorization of features in these teacher model is one of the main reasons for improved accuracy in the student model.
 
 In our attacks against word-based embedding  models, we fine-tune the \glove{} embedding layer with the same learning rate as the student model. However, fine-tuning does not help in mitigating these attacks as our adversarial examples do not rely on 
\glove{} vectors but the fact that the student model uses certain words as important features. Table~\ref{tab:model_acc} gives the attack accuracy for both the \glove$_{\tt FE}$ and \glove$_{\tt FT}$. However, our sentence based adversarial attack is based on the BERT output vector. Hence, retraining the initial layers make the attack difficult to succeed as the output of the BERT layer changes as compared to the public pre-trained model. When we fine-tune the \bert{} model with a higher learning rate of 0.001 for the Fake News Detection task, we observe that fine-tuning drops the accuracy of the student model i.e., it reduces from $92\%$ to  $70\%$. This observation is in alignment with the results from prior work~\cite{yao2019latent}.

\subsection{Dropout}


Dropouts in deep learning model serve the purpose of making the model more generalizable. This makes sure that the model is not overfitted to the training data set. Therefore, dropouts are shown to be good at preventing the model from memorizing the training dataset~\cite{srivastava2014dropout}. We introduce a dropout ratio of $0.5$ inside and after the LSTM layer in our binary classification tasks making it generalizable and achieving better test accuracy. Results in Section \ref{sec:wembedding_res} show that after adding dropouts the attack accuracy improves. Dropouts do not mitigate our attacks since they do not rely on the over-fitting of the model. Our attacks leverage any unintended feature that the model learns for prediction of the task while dropouts do not ensure unwanted learning.
 However, increasing the dropout ratio may prevent the model from making predictions based on words but will reduce the test accuracy as the model does not learn the right features to make correct predictions.

\subsection{Adversarial Training}

Adversarial training refers to the process of adding adversarial examples to the training dataset. This makes the model robust against those adversarial examples. Such behavior is expected as the model now learns that these features are not intended for the given classification task. However, the models still remain vulnerable to other adversarial examples that are generated using some other unintended features that the model learns for prediction. Alzantot et al. perform a similar attack to ours using word-based features for generating adversarial samples albeit using a different attack strategy~\cite{alzantot2018generating}. Their attack algorithm is $100\%$ robust towards adversarial training. This is because their algorithm finds some other words that are vulnerable and not included in the training dataset. Our word-score based attack creates adversarial examples using words with highest word-score.  Making the model robust against a few words does not prevent the algorithm from selecting other words as unintended features for prediction. Thus, adversarial training is not a robust defense against our attack.

 \section{Discussion}

\label{sec:discussion}

Our attacks exploit the fact that models do not always learn and use intended features to predict the output.  Our adversarial examples are successful mainly due to the vulnerability of the target models to unintended features.
The basic idea of transfer learning is to learn from a model trained on large data set and thereby contains various unintended features by design. This makes transfer-learned models more susceptible to these attacks exacerbated by the fact that an attacker has access to these public features. This raises the question: \emph{whether we should use a transfer-learned model to achieve better accuracy at the expense of security risk?} 

Moreover, existing defenses are not robust enough to prevent these attacks without compromising model accuracy.
Our results stress on the importance of making text classification models robust against malicious inputs. While data augmentation based solutions like adversarial training have shown to reduce the attack accuracy, adding all valid perturbations without reducing the model accuracy is an open question. Another approach is to make the loss function objective robust against perturbations as well instead of just optimizing the accuracy. All these above solutions raise the question of properly specifying valid perturbations for a given model. 

  \section{Related Work}
\label{sec:related}

Transfer learning is shown to be effective in different domains. Yosinski et al. \cite{yosinski2014transferable} compared various transfer learning approaches and examined their impact on model performance for images. Transfer learning techniques were also shown to be beneficial in improving accuracy on text classification models and reducing the training time \cite{howard2018universal}. But the question is: \emph{Is transfer learning from public teacher model a good decision?} Not much research has been done to understand the limitations of using public models for transfer learning. 

\paragraph{Attacks on transfer learning for images.}
Shafahi et al.\cite{shafahi2018poison} were the first to propose attacks in transfer-learned models. They showed how to poison a student model, by crafting poisoned training datasets based on features of the teacher model. Recently, Wang et al. and Yao et al. demonstrated adversarial examples and latent backdoor attacks on transfer learned image recognition models~\cite{wang2018great,yao2019latent}. However, these few attacks that exploit the public teacher model only consider image-based inputs. For text data, there has been work on adversarial attacks only on the non-transfer learned setting.

\paragraph{Attacks on Non-transfer Learned  Text  Models.}
Papernot el al. \cite{papernot2016crafting} first explored generating adversarial attacks for text data based on fast gradient sign method (FGSM). Prior work on generating adversarial text mostly proposed various heuristics to optimize the objective function by replacing few words or characters with similar alternatives that result in misclassification of the input. The focus was on identifying the important features (mostly words) that led to the classification decision and then trying to replace it with some other feature that classifies the input differently. Some common approaches to identify the important words are removing the word from the sentence and checking the difference in confidence  \cite{lei2019discrete}  or replacing it white spaces \cite{gao2018deep}. The white-box based methods identified important words using gradient-based approach like FGSM assuming they have access to model weights \cite{lei2019discrete},\cite{sato2018interpretable},\cite{gong2018adversarial},\cite{samanta2017towards}. The important words are replaced to unknown words by introducing a spelling error \cite{liang2017deep},\cite{li2018textbugger} or replacing them by synonyms \cite{alzantot2018generating},\cite{lei2019discrete}. Both these techniques are based on trial and error and do not guarantee that the input will get misclassified. Moreover, they require several tries i.e., proportional to the number of words in the input to perform the attack successfully for every input. Our adversarial examples are robust against common defenses like trivial spell checkers and required limited queries to the student model.

  \section{Conclusion}

We present the first attack algorithms for generating adversarial inputs for text-classification tasks in a transfer-learning setting. 
We perform attacks for two types of teacher models: word-level and sentence-level based text embedding models. We consider the word-based model, where we exploit the fact that the model is biased to the presence of certain words for certain classes. 
 We also perform the first attack against sentence-based embedding (\bert) using unintended features like the length of sentence. We demonstrate that traditional defenses are ineffective against our attack. Mitigating our attacks require techniques that prevent the model from using unintended features for prediction. However, we speculate that enforcing such constraints might degrade the accuracy of the model which is not acceptable in practice.  

\bibliographystyle{IEEEtranS}
\bibliography{paper}

\end{document}